\documentclass[final,3p,times]{elsarticle}

\usepackage{amssymb}
\usepackage{amsmath, amsfonts}

\usepackage{microtype}      
\usepackage{xcolor}         
\usepackage{multirow}
\usepackage{makecell}
\usepackage[table]{xcolor}
\usepackage{hyperref}
\usepackage{cleveref}
\crefname{figure}{Fig.}{Figs.}
\crefname{table}{Table}{Tables}
\crefname{section}{Section}{Sections}
\usepackage{graphicx}    
\usepackage{subcaption}     
\usepackage{float}
\usepackage{algorithmic}
\usepackage{algorithm}
\usepackage{booktabs}
\usepackage{cleveref}
\crefname{figure}{Fig.}{Figs.}
\crefname{table}{Table}{Tables}
\crefname{equation}{Eq.}{Eqs.}
\crefname{section}{Section}{Sections}
\usepackage{hyperref}

\journal{Pattern Recognition}

\begin{document}

\begin{frontmatter}



\title{Perceptual Anchoring: Prototype-Guided Text Calibration for Training-free Open-Vocabulary Semantic Segmentation}

\author[1,2]{Wanli Ma}
\author[3]{Jiangwen Lu}
\author[1,2]{Qinmu Peng \corref{cor1}}
\author[1,2]{Xinge You}

\cortext[cor1]{Corresponding author}

\address[1]{School of Electronic Information and Communications, Huazhong University of Science and Technology, Wuhan 430074, China}
\address[2]{National Anti-counterfeit Engineering Research Center, Huazhong University of Science and Technology, Wuhan 430074, China}
\address[3]{School of Automation, Hangzhou Dianzi University, Hangzhou 310018, China}

\begin{abstract}
Training-free open-vocabulary semantic segmentation (OVSS) partitions an image into semantically distinct regions based on arbitrary text descriptions, without learning any additional parameters. However, existing methods typically focus on improving visual representations while treating text embeddings that encode only generic category concepts as fixed classification references. The resulting semantic gap between these generic concepts and the visual representations that capture the specific appearances of target instances often causes incomplete masks and erroneous predictions in non-target regions. Inspired by the symbol-percept correspondence underlying perceptual anchoring, we propose Prototype-Guided Text Calibration (PTC) for training-free OVSS. In the Perceiving stage, PTC selects reliable visual evidence based on initial matching scores to construct category-specific visual prototypes. In the Anchoring stage, PTC uses these prototypes to calibrate their corresponding text embeddings, with the calibration strength adaptively adjusted based on the amount of visual evidence. Consequently, the calibrated text embeddings align more accurately with instance-specific visual representations while preserving generic category semantics and open-vocabulary generalization. Moreover, PTC requires neither additional training nor external models and can serve as a plug-and-play module for existing methods. Extensive experiments across eight benchmarks show that PTC significantly enhances the performance of six representative methods and yields more complete and accurate segmentation results. These results validate PTC as a simple and effective approach to improving visual-text alignment.
The code is available at [\url{https://github.com/Valeria-MaW/PTC}].
\end{abstract}



\end{frontmatter}


\section{Introduction}
\label{sec1}
Open-vocabulary semantic segmentation (OVSS) aims to partition an image into semantically distinct regions using arbitrary text descriptions. 
Unlike traditional semantic segmentation methods that rely on closed-set categories and costly dense annotations \citep{Chen2018Deeplab, Xie2021SegFormer, Cheng2022Mask2Transformer, Guo2022SegNeXt}, OVSS can recognize classes unseen during training, making it better suited to real-world visual perception, which encompasses tens of thousands of categories.

Large-scale vision-language pre-trained models (VLMs) (e.g., CLIP \citep{Radford2021CLIP} and ALIGN \citep{Jia2021ALIGN}) have demonstrated impressive transferability in recognizing novel classes. 
This prompts researchers to apply them to OVSS and to explore their adaptation to dense prediction tasks \citep{Ding2022ZegFormer, Jiao2023MAFT1, Xu2022GroupViT, Xing2023CoCu, Wang2024SCLIP, Lan2024proxyclip}, since CLIP is inherently designed for image-level visual-text matching.
Existing OVSS methods can be broadly divided into training-based and training-free approaches, depending on whether additional parameter learning is required. Training-based approaches typically rely on large segmentation datasets \citep{Ding2022ZegFormer, Jiao2023MAFT1, Jiao2024MAFT2, Han2023GKC}, image-text pair datasets \citep{Xu2022GroupViT, Xing2023CoCu, Ranasinghe2023CLIPpy}, or other forms of supervision \citep{Chen2023ZeroSeg, Wu2024CLIPSelf}, introducing significant computational costs. 
In contrast, training-free OVSS has attracted increasing attention because it requires no additional parameter learning and facilitates generalization to new scenes and classes.

\begin{figure}[t]
\centering
\includegraphics[width=\linewidth]{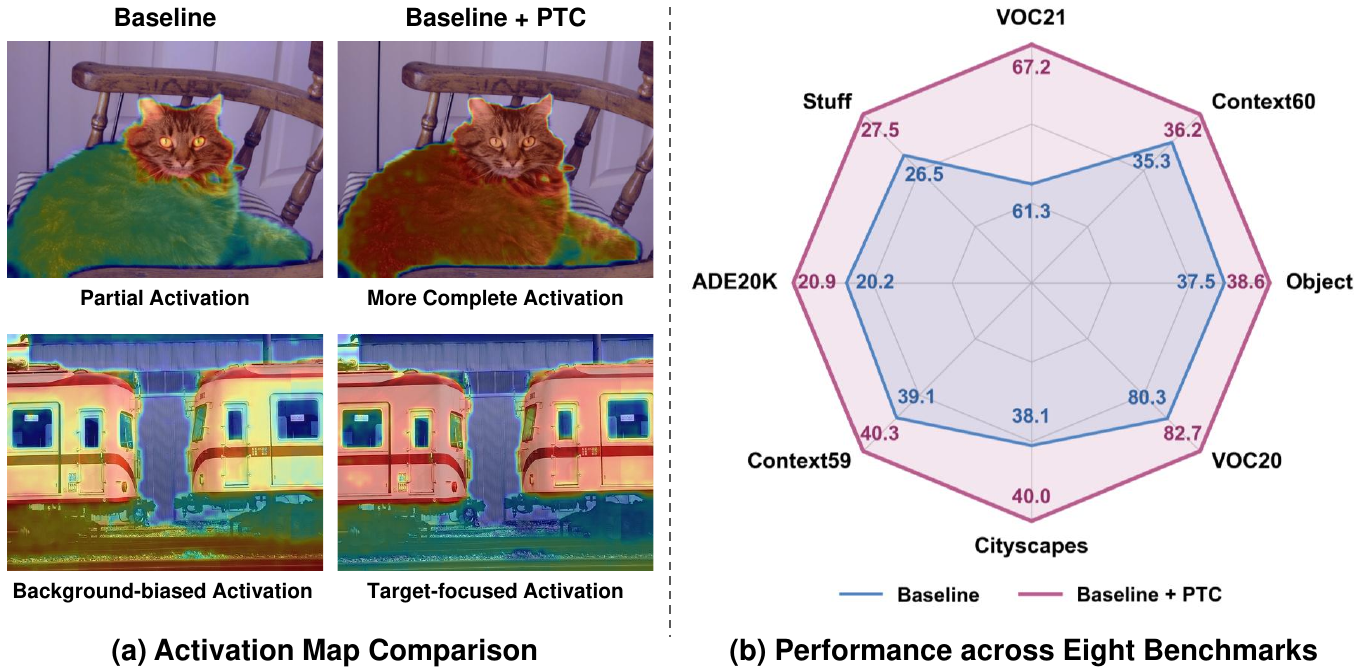}
\caption{\textbf{Motivation and effectiveness of PTC}. 
\textbf{(a)} Compared with the ProxyCLIP baseline, adding PTC yields more complete activation over object regions and suppresses background-biased responses.
\textbf{(b)} Introducing PTC consistently improves ProxyCLIP across all eight benchmarks.
}
\label{fig1}
\end{figure}

Most existing training-free methods focus on improving visual representations, primarily by enhancing their local perception and spatial consistency, thereby improving visual–text matching and segmentation performance \citep{Zhou2022MaskCLIP, Wang2024SCLIP, Hajimiri2025NACLIP, LI2025CLIP_Surgery, Bousselham2024GEM, Shao2025CLIPTrase, Lan2024ClearCLIP, Yang2025ResCLIP, Lan2024proxyclip, Zhang2025CorrCLIP, kirillov2023sam, Caron2021Dino, Shi2025Trident}. However, these methods often treat class text embeddings as reliable zero-shot classifiers and directly match them to visual features, which overlooks a semantic gap between category text embeddings and visual representations.
Specifically, class text embeddings generated by pre-trained large language models (LLMs) \citep{radford2019GPT-2, lee2018BERT} learn generic, input-agnostic category concepts. In contrast, visual representations are extracted from the input image and characterize the specific appearances of target instances under particular poses, viewpoints, or imaging conditions.
Consequently, directly matching generic category text embeddings with instance-specific visual representations inevitably suffers from semantic misalignment, which naturally degrades segmentation performance and manifests in two limitations. First, the model tends to recognize only local regions that are highly representative of the category, resulting in incomplete mask coverage, as shown in the upper part of \cref{fig1}(a). Second, generic text embeddings may be incorrectly matched with visually similar background regions or with regions belonging to co-occurring classes, resulting in erroneous segmentation, as shown in the lower part of \cref{fig1}(a).

Since final segmentation predictions inherently rely on visual–text matching, refining only visual representations while leaving generic text embeddings unchanged fails to fully mitigate the aforementioned semantic gap.
To bridge this gap, we draw inspiration from perceptual anchoring in cognitive robotics, a conceptual mechanism that links abstract symbols to concrete percepts, allowing the symbols to refer to specific entities \citep{coradeschi2001perceptual_anchoring, coradeschi2003introduction_anchoring}.
From this perspective, we argue that category text embeddings should not always be treated as fixed classification references.
Instead, they should be linked to the instance-specific visual representations in the input image using reliable perceptual evidence, enabling these abstract category concepts to more accurately refer to the target instances, thus achieving more precise segmentation.

Building on this insight, we propose Prototype-Guided Text Calibration (PTC) for training-free OVSS.
In the Perceiving stage, PTC performs a reliability assessment and an evidence-quantity strategy to mitigate noise interference and to accommodate target-scale variations, thereby selecting reliable visual evidence based on the initial matching scores to construct category prototypes. In the Anchoring stage, PTC uses these prototypes to calibrate the corresponding text embeddings and adaptively adjusts the calibration strength based on evidence sufficiency to prevent semantic deviations caused by over-reliance on limited evidence.
In this way, we establish a link between generic category concepts and category instances with specific appearances, which enables calibrated text embeddings to preserve generic semantics while aligning more closely with instance-specific visual representations, thereby improving final segmentation performance.
The main contributions of this work are as follows:

\begin{itemize}
\item We identify an underexplored semantic gap in training-free OVSS between generic category text embeddings and instance-specific visual representations derived from the input, which limits the accuracy of visual-text alignment and thereby constrains segmentation performance.

\item We propose PTC, a novel prototype-guided text calibration method that can be integrated into various training-free OVSS methods as a plug-and-play module. It calibrates category text embeddings using visual prototypes with evidence-driven adaptive calibration, aligning them more closely with visual representations while preserving generic semantics, thereby improving visual–text matching. 

\item We conduct extensive experiments on eight segmentation benchmarks and integrate PTC into six representative training-free OVSS baselines. Quantitative and qualitative results validate the effectiveness of PTC.
\end{itemize}

\section{Related Work}
\label{sec2}
\paragraph{\textbf{Visual representation enhancement}} Most existing training-free OVSS methods directly modify the internal architecture of CLIP, enabling its visual representations to capture local information.
MaskCLIP \citep{Zhou2022MaskCLIP} modifies the final-layer self-attention to exploit CLIP's latent localization capability. SCLIP \citep{Wang2024SCLIP} replaces this with correlative self-attention to enhance spatial consistency of visual representations. NACLIP \citep{Hajimiri2025NACLIP} extends this by strengthening local neighborhood modeling to improve the continuity of predictions across neighboring regions. CLIP Surgery \citep{LI2025CLIP_Surgery}, GEM \citep{Bousselham2024GEM}, and CLIPtrase \citep{Shao2025CLIPTrase} introduce more flexible and generalized formulations of self-attention, thereby enhancing segmentation performance. ClearCLIP \citep{Lan2024ClearCLIP} removes the residual branch and feed-forward network from the final layer to obtain cleaner dense representations. ResCLIP \citep{Yang2025ResCLIP} exploits localization information in intermediate-layer attention to improve category consistency.
Moreover, other studies leverage spatial information from external models to compensate for CLIP’s limited local perception.
ProxyCLIP \citep{Lan2024proxyclip} constructs proxy attention from spatial correspondences between vision foundation models (VFMs), e.g., SAM \citep{kirillov2023sam}, DINO \citep{Caron2021Dino}, DINOv2 \citep{oquab2023dinov2}, and CLIP features to enhance CLIP's local consistency. CorrCLIP \citep{Zhang2025CorrCLIP} uses SAM to constrain patch interactions and reconstructs more coherent patch correlations by self-supervised vision models, thereby improving region predictions. For high-resolution images, Trident \citep{Shi2025Trident} proposes a splice-then-segment paradigm, which first splices CLIP and DINO features from sub-images, then uses the global correlations provided by SAM to aggregate features and refine predictions.
Despite these advancements, these methods typically treat category text embeddings as fixed classification references and overlook the inherent semantic gap between generic category text embeddings and the instance-specific visual representations in the input, leading to limited segmentation performance.

\paragraph{\textbf{Visual reference construction}} Recent efforts use external models to construct visual references and segment target objects based on robust correspondences in the visual representation space.
FreeDA \citep{Barsellotti2024FreeDA} constructs visual references from images generated by Stable Diffusion \citep{rombach2022diffusion} and performs prediction by combining local region-to-prototype similarity with global image-to-text similarity. FOSSIL \citep{Barsellotti2024FOSSIL} constructs retrievable synthetic visual references and uses OpenCut to generate candidate regions from input images, then matches their embeddings with prototypes in a self-supervised visual space. Moreover, RIM \citep{Wang2024RIM} uses Stable Diffusion \citep{rombach2022diffusion} and SAM to construct image references and to perform image-to-image matching via relation-aware ranking in the DINOv2 feature space.
Rather than directly improving visual features, these methods construct category-level visual references for visual-visual matching, alleviating the ambiguity and noise caused by the modality gap during visual-text matching.
Notably, these references represent generic visual patterns for each category as they are externally constructed from other pre-trained models.
Consequently, these methods still essentially match generic semantics with instance-specific representations, and while useful for target localization, they struggle to accurately align object boundaries for dense prediction.
Instead, PTC directly constructs visual prototypes from reliable visual evidence in the current input, which are not used to replace text embeddings but to calibrate the original category text embeddings, aiming to establish a perceptual link between generic concepts and the specific visual appearances of the corresponding instances. This enables the calibrated text embeddings to align more closely in semantics with instance-specific visual representations, thereby mitigating the semantic gap.

\paragraph{\textbf{Text-side adaptation}} Several studies explore how category text descriptions affect training-free OVSS.
LLM-Supervision \citep{sun2024LLM_Supervision} aggregates predictions from GPT-3 \citep{brown2020GPT-3}-generated subclass descriptions to represent the segmentation mask of the target category, whereas ITACLIP \citep{Aydin2025ITACLIP} directly fuses text embeddings of category synonyms for prediction. 
Moreover, FreeCP \citep{Chen2025FreeCP} purifies the candidate category set to reduce the effects of category redundancy and confusion on segmentation results. 
Although the first two methods enrich category descriptions, they still rely on fixed, pre-generated text that encodes only generic semantics. Meanwhile, FreeCP primarily focuses on filtering candidate classes to mitigate confusion. Consequently, the semantic gap between generic category text embeddings and instance-specific visual representations, along with the resulting limited segmentation performance, remains underexplored.


\section{Methodology}
\label{sec3}
\subsection{Preliminaries}
\label{sec:3.1}
Given an input image $I$ and a target vocabulary $\mathcal{Y}=\{y_1, y_2, \dots, y_C\}$, training-free OVSS aims to assign each pixel in the image a semantic label from $\mathcal{Y}$ without any additional training. 

\paragraph{\textbf{CLIP-based Segmentor}} 
In the CLIP visual encoder, which is built on the Vision Transformer \citep{Vaswani2017ViT}, the input image $I$ is first mapped to visual features $F=[f_1,f_2,\dots,f_N]^\top\in\mathbb{R}^{N\times D}$, where $N$ is the number of tokens and $f_i\in\mathbb{R}^{D}$ denotes the $D$-dimensional feature of the i-th token.
Each category name $y_c \in \mathcal{Y}$ is combined with a predefined prompt template (e.g., “a photo of a \{class name\}”) and encoded by the text encoder to produce the corresponding class text embedding $t_c \in\mathbb{R}^{D}$. The resulting text embeddings are stacked as $T=[t_1,t_2,\dots,t_C]^\top\in\mathbb{R}^{C\times D}$. 
Then, the model computes the cosine similarities between the visual features and category text embeddings to obtain the initial matching scores:
\begin{equation}
S= \cos(F, T)
\label{initialS}
\end{equation}
where $S_{i,c} = \cos(f_i, t_c) = \frac{f_i^\top t_c}{\|f_i\|_2 \|t_c\|_2}$ denotes the matching score between the $i$-th visual token and the $c$-th text embedding. Finally, the segmentation map is obtained as
\begin{equation}
\mathcal{M}=\operatorname*{arg\,max}_{c} S
\end{equation}

Although most existing methods improve segmentation performance by enhancing the ability of visual features to capture local information, they generally treat the category text embeddings produced by the pre-trained text encoder as fixed classification references.
However, class text embeddings primarily encode generic concepts learned through large-scale image–text pre-training, while visual representations are extracted directly from the input image to characterize the specific visual appearances of the target object in the current scene.
Therefore, the process in \cref{initialS} amounts to directly matching generic category text embeddings with instance-specific visual representations, which inevitably suffers from visual-text misalignment due to the semantic gap, thus resulting in incomplete or erroneous predictions even for a strong baseline, as illustrated in \cref{fig1}(a).
Inspired by perceptual anchoring, we argue that narrowing this semantic gap requires establishing a perceptual link between generic category concepts and the corresponding target instances in the input image. Accordingly, category text embeddings should not remain standalone classification references, but should instead be linked to instance-specific visual representations.
Based on this insight, we propose PTC, whose details are presented in \cref{sec3.2}.

\begin{figure*}[t]
\centering
\includegraphics[width=\textwidth]{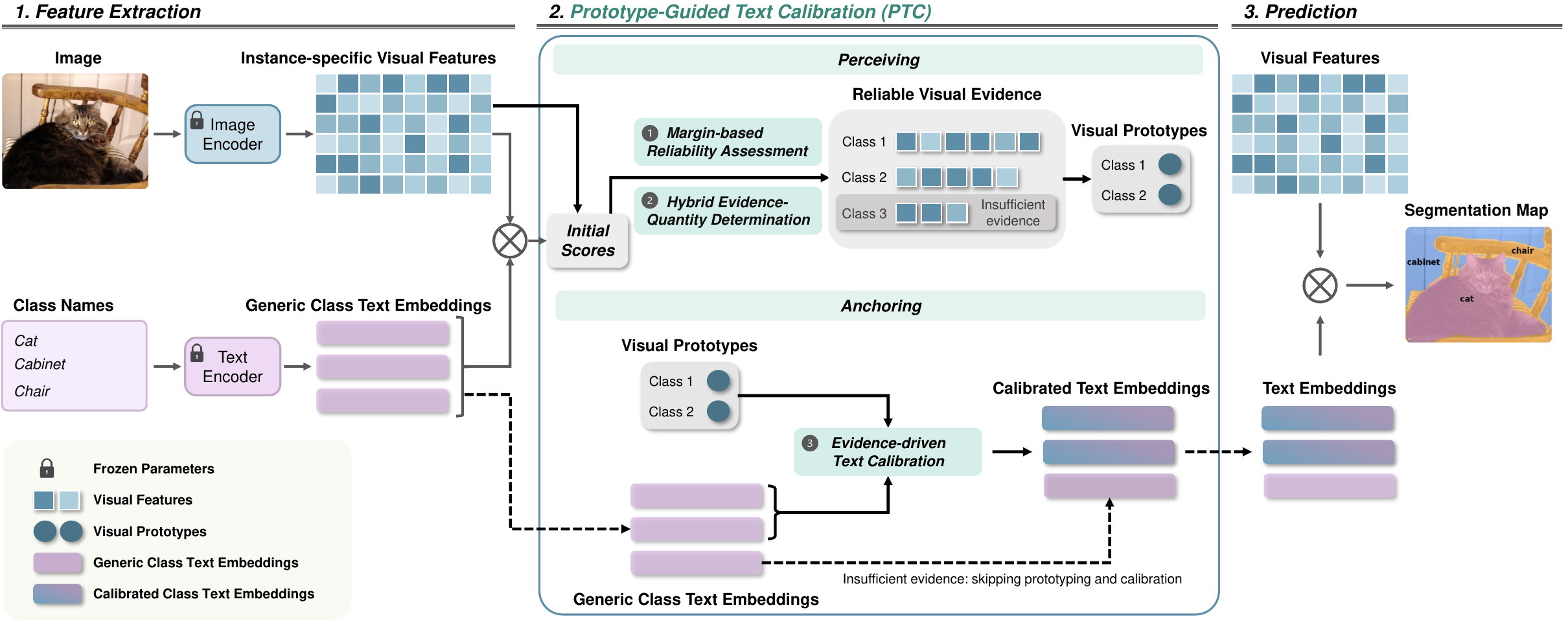}
\caption{
\textbf{Overall framework of PTC for training-free OVSS.} 
\textbf{1. Feature Extraction:} The input image and class names are encoded by frozen encoders to obtain visual features and text embeddings, respectively. 
\textbf{2. PTC:} In the \textit{Perceiving} stage, reliable visual evidence is selected from the visual features based on initial matching scores and aggregated into visual prototypes. In the \textit{Anchoring} stage, the text embeddings of categories with sufficient visual evidence are calibrated using their corresponding prototypes, while those with insufficient visual evidence remain unchanged. 
\textbf{3. Prediction:} The resulting text embeddings are matched with the visual features to generate the final segmentation map.}
\label{fig2}
\end{figure*}

\subsection{Prototype-Guided Text Calibration}
\label{sec3.2}
\subsubsection{Overview}
To mitigate the semantic gap between generic category text embeddings and instance-specific visual representations in the input image, we propose Prototype-Guided Text Calibration (PTC). 
It aims to use reliable visual evidence from the image to establish a perceptual link between text and visual representations, allowing class text embeddings to incorporate the specific visual cues of category instances while preserving generic semantics, thereby enabling more accurate visual–text alignment. 
The overall framework of PTC is shown in \cref{fig2}.
Specifically, PTC comprises two core stages: Perceiving and Anchoring. 
In the Perceiving stage, PTC extracts reliable visual evidence and aggregates it into category-specific visual prototypes that capture the appearances of the corresponding category instances in the input image.
In the Anchoring stage, PTC uses these prototypes to calibrate the original category text embeddings. For categories with insufficient visual evidence, PTC skips prototype construction and text calibration, thereby preventing semantic deviations caused by over-reliance on limited evidence.

We denote the calibrated text embeddings by
$T_{\mathrm{cal}}$, and the final visual-text matching scores and segmentation map are then computed as
\begin{equation}
S_{\mathrm{new}}=\cos(F,T_{\mathrm{cal}}), \qquad
\mathcal{M}_{\mathrm{new}}=\operatorname*{arg\,max}_{c} S_{\mathrm{new}}
\label{FinalS}
\end{equation}

Since initial predictions are often affected by background noise and category confusion, constructing visual prototypes directly from the corresponding visual features may introduce unreliable information, thereby compromising the generic category semantics in the original text embeddings during subsequent calibration.
Consequently, PTC addresses two key design challenges.
The first is to extract reliable visual evidence from noisy initial matching scores and determine an appropriate evidence quantity across varying object scales, thereby constructing robust visual prototypes (\cref{sec3.2.2}).
The second is to incorporate instance-specific visual information into the text embeddings while preserving their original generic semantics (\cref{sec3.2.3}).

\subsubsection{Perceiving: Reliable Visual Prototype Construction}
\label{sec3.2.2}
In the Perceiving stage, PTC extracts reliable visual evidence from visual features and aggregates it into class-specific visual prototypes that characterize the appearance of the category instance in the current image. This stage inherently faces two key challenges.
First, the initial matching scores are susceptible to background noise and category confusion, making it essential to evaluate the reliability of candidate tokens to construct high-quality visual prototypes.
Second, it is challenging to adapt the number of evidence tokens to variations in object scale, which requires a careful balance between ensuring sufficient tokens to guarantee prototype representativeness and avoiding the oversampling of noisy or ambiguous tokens that could contaminate the prototype.
To overcome these challenges, PTC measures token reliability using the score margin and employs a hybrid strategy to determine the number of evidence tokens required for each category.

Given the initial matching scores $S$ computed by \cref{initialS}, the initially predicted category for the $i$-th visual token is $\hat{c}_i=\arg\max_c S_{i,c}$. Thus, the candidate token set of category $c$ is defined as

\begin{equation}
\Omega_c = \{i \mid \hat{c}_i=c\}, \qquad
N_c = |\Omega_c|
\end{equation}
where $N_c$ denotes the number of visual tokens initially predicted as category $c$.

\paragraph{\textbf{Margin-based Reliability Assessment}}
To select reliable visual tokens from $\Omega_c$ as visual evidence, we introduce the score margin to quantify the certainty of the predicted category for each candidate token:

\begin{equation}
\Delta_i = S_{i,\hat{c}_i} - \max_{c'\neq\hat{c}_i} S_{i,c'}
\end{equation}
where $\Delta_i$ denotes the difference between the matching scores of the category with the highest score for the $i$-th token and the category with the second-highest score. 
A larger $\Delta_i$ indicates a greater score advantage of the predicted category over its strongest competitor, meaning the token's category assignment is less ambiguous and thus more reliable as visual evidence.
Compared with the absolute confidence that reflects only the response strength for a single class, the score margin is a relative measure that can effectively filter out ambiguous tokens that have high absolute confidence in the predicted category but also respond strongly to other categories, thereby ensuring reliable and stable prototype construction.

\paragraph{\textbf{Hybrid Evidence-Quantity Determination}}
To account for differences in spatial scales and candidate set sizes across categories, we design a hybrid strategy that combines a minimum-count constraint with proportional sampling to determine the target number of tokens selected as visual evidence for category $c$:

\begin{equation}
K_c = \min\left(N_c,\max\left(K_{\min},\left\lfloor \rho N_c \right\rfloor\right)\right)
\label{eq:evidence_quantity}
\end{equation}
where $K_{\min}$ denotes the minimum number of tokens required to construct a reliable visual prototype, while $\rho$ is the sampling ratio and is set to $0.1$ by default.

The hybrid strategy is designed to provide sufficient visual evidence across different categories with various candidate set sizes.
When the candidate set is relatively small, the $K_{\min}$ term prevents the prototype from exhibiting poor representativeness due to being constructed from only a few tokens.
When the candidate set is large, the proportional term allows more visual evidence to be collected, providing a more complete characterization of the instance appearance.
Meanwhile, $N_c$ serves as an upper bound, ensuring that the actual number of selected tokens does not exceed the number of candidate tokens available for the category.
Categories with $N_c<K_{\min}$ are regarded as having insufficient evidence and are excluded from subsequent prototype construction and text calibration.

Then, PTC selects the $K_c$ tokens with the largest score margins from the candidate set $\Omega_c$ as visual evidence:
\begin{equation}
\mathcal{E}_c = \operatorname{TopK}_{i\in\Omega_c} \left( \Delta_i, K_c \right)
\end{equation}
where $\mathcal{E}_c$ denotes the index set of the $K_c$ selected evidence tokens for category $c$.

Subsequently, PTC aggregates the selected tokens using their score margins as weights:
\begin{equation}
V_c^{\mathrm{proto}} =  \sum_{i\in\mathcal{E}_c} w_i f_i, \qquad
w_i = \frac{\Delta_i}{ \sum_{j\in\mathcal{E}_c}\Delta_j+\epsilon}
\end{equation}
where $V_c^{\mathrm{proto}}$ is the visual prototype for category $c$, $w_i$ denotes the weight assigned to the $i$-th visual token during prototype aggregation, and $\epsilon$ is a small constant used to prevent division by zero.
This weighting scheme assigns larger weights to tokens with less ambiguous category assignments while reducing the influence of low-margin tokens, thereby improving prototype stability.

\subsubsection{Anchoring: Evidence-Driven Text Calibration}
\label{sec3.2.3}
In the Anchoring stage, PTC uses category visual prototypes to calibrate the corresponding text embeddings, incorporating specific visual cues from category instances while preserving their generic category semantics.
Achieving this goal requires appropriately controlling the influence of the visual prototypes on the text embeddings, i.e., the calibration strength.
Since each prototype is formed by the weighted aggregation of visual evidence selected in the Perceiving stage, the number of evidence tokens can indicate whether the prototype is sufficiently supported.
A larger evidence set provides broader support for the prototype and reduces its sensitivity to any individual token. In contrast, with limited evidence, an erroneous token can exert a greater influence on the aggregated prototype.
Accordingly, PTC determines whether each category is eligible for calibration and adaptively adjusts its calibration strength based on the sufficiency of its visual evidence.

First, the set of categories eligible for calibration is determined based on the amount of visual evidence:
\begin{equation}
\mathcal{C}_{\mathrm{valid}} = \left\{ c \mid n_c^{\mathrm{ev}} \geq K_{\min} \right\}
\end{equation}
where $n_c^{\mathrm{ev}}=|\mathcal{E}_c|$ denotes the amount of visual evidence for category $c$.
For classes with $n_c^{\mathrm{ev}}<K_{\min}$, PTC leaves the corresponding text embeddings unchanged to avoid semantic deviations caused by calibration with limited evidence.
For categories that meet the minimum evidence threshold, PTC further adjusts the calibration strength according to the amount of visual evidence.

Specifically, for each $c\in\mathcal{C}_{\mathrm{valid}}$, its calibration strength $\mu_c$ is computed as:
\begin{equation}
\mu_c = \mu\cdot\alpha_c, \quad
\alpha_c = \min\left(1, \frac{\log\left(1+n_c^{\mathrm{ev}}\right)}{\log\left(1+\lambda K_{\min}\right)}\right)
\label{eq:calibration}
\end{equation}
where $\mu$ denotes the global calibration strength and also serves as the upper bound of the category calibration strength $\mu_c$, $\alpha_c$ is the adaptive calibration coefficient for category $c$, and $\lambda$ controls the amount of evidence required for the category calibration strength to reach its upper bound.
A larger $\lambda$ raises the amount of evidence required for $\mu_c$ to reach $\mu$.
When $K_{\min}\leq n_c^{\mathrm{ev}}<\lambda K_{\min}$, $\mu_c$ increases logarithmically with the number of evidence tokens. When $n_c^{\mathrm{ev}}\geq\lambda K_{\min}$, $\mu_c$ saturates at $\mu$.
Therefore, categories that meet the threshold but have relatively limited evidence are calibrated conservatively.
Meanwhile, the calibration strength no longer increases with the evidence count once it becomes sufficient, preventing text embeddings associated with large objects from deviating from their original category semantics due to progressively stronger calibration solely because they provide more visual evidence.

Then, category text embeddings are updated as follows:
\begin{equation}
t_c^{\mathrm{cal}} = \begin{cases}  (1-\mu_c)t_c + \mu_c V_c^{\mathrm{proto}} , & c\in\mathcal{C}_{\mathrm{valid}} \\
t_c, & c\notin\mathcal{C}_{\mathrm{valid}}
\end{cases}
\end{equation}

By calibrating based on the sufficiency of visual evidence, PTC allows text embeddings to incorporate an appropriate amount of category-specific visual information from the input image while preserving their original generic category semantics, thus improving visual-text matching.
The final matching scores and the segmentation map are obtained as specified in \cref{FinalS}.

\section{Experiments}
\label{sec4}
\subsection{Implementation Details}
\label{sec:4.1}
\paragraph{\textbf{Datasets}}
We conduct a comprehensive evaluation on eight well-established datasets and categorize these datasets into two groups: 
(1) \textbf{with the background class}: PASCAL VOC21 (VOC21) \citep{Everingham2015VOC}, PASCAL Context60 (Context60) \citep{2014Context}, COCO-Object (Object) \citep{Caesar2018StuffObject};
(2) \textbf{without the background class}: PASCAL VOC20 (VOC20) \citep{Everingham2015VOC}, Cityscapes (City) \citep{2016City}, PASCAL Context59 (Context59) \citep{2014Context}, ADE20K-150 (ADE) \citep{Zhou2019ade}, COCO-Stuff (Stuff) \citep{Caesar2018StuffObject}. 
All experiments strictly followed the original dataset division guidelines to ensure the reliability of the assessment results.

\begin{table*}[t]
\centering
\caption{\textbf{Comparison with state-of-the-art training-free methods on eight benchmarks.}
The "External Models" column indicates whether the corresponding method relies on external models, such as VFMs or LLMs.
\textcolor{blue}{Blue values} denote the absolute improvements achieved by adding PTC to the corresponding baseline.
Results with $^{*}$ are postprocessed with DenseCRF.}
\label{tab:comparison}
\resizebox{\textwidth}{!}{
\begin{tabular}{l l lll lllll l}
\toprule
\multirow{2}{*}{Methods}
& \multirow{2}{*}{\shortstack{External\\Models}}
& \multicolumn{3}{c}{w background}
& \multicolumn{5}{c}{w/o background}
& \multirow{2}{*}{Avg.} \\
\cmidrule(lr){3-5} \cmidrule(lr){6-10}
& & VOC21 & Context60 & Object & VOC20 & City & Context59 & ADE & Stuff & \\
\midrule

CLIP {\scriptsize \textcolor{gray}{[ICML'21]}} \citep{Radford2021CLIP}
& $\times$ & 18.8 & 9.9 & 8.1 & 49.4 & 6.5 & 11.1 & 3.1 & 5.7 & 14.1 \\

ReCo {\scriptsize \textcolor{gray}{[CVPR'22]}} \citep{Shin2022ReCo}
& $\times$ & 25.1 & 19.9 & 15.7 & 57.7 & 21.6 & 22.3 & 11.2 & 14.8 & 23.5 \\

MaskCLIP {\scriptsize \textcolor{gray}{[ECCV'22]}} \citep{Zhou2022MaskCLIP}
& $\times$ & 43.4 & 23.2 & 20.6 & 74.9 & 24.9 & 26.4 & 11.9 & 16.7 & 30.3 \\

CaR$^{*}$ {\scriptsize \textcolor{gray}{[CVPR'24]}} \citep{Sun2024CaR}
& $\times$ & 67.6 & 30.5 & 36.6 & 91.4 & - & 39.5 & 17.7 & - & - \\

GEM {\scriptsize \textcolor{gray}{[CVPR'24]}} \citep{Bousselham2024GEM}
& $\times$ & 56.9 & 32.6 & 31.1 & 79.9 & 30.8 & 35.9 & 15.7 & 23.7 & 38.3 \\

CLIPtrase {\scriptsize \textcolor{gray}{[ECCV'24]}} \citep{Shao2025CLIPTrase}
& $\times$ & 53.0 & 30.8 & 44.8 & 81.2 & - & 34.9 & 17.0 & 24.1 & - \\

OVDiff {\scriptsize \textcolor{gray}{[ECCV'24]}} \citep{Karazija2025OVDiff}
& $\times$ & 62.8 & 28.6 & 34.6 & 80.9 & 23.4 & 32.9 & 14.1 & 20.3 & - \\

SCLIP {\scriptsize \textcolor{gray}{[ECCV'24]}} \citep{Wang2024SCLIP}
& $\times$ & 59.1 & 30.4 & 30.5 & 80.4 & 32.2 & 34.2 & 16.1 & 22.4 & 38.2 \\

SCLIP + PTC {\scriptsize [Ours]}
& $\times$
& 61.5{\scriptsize\textcolor{blue}{(+2.4)}}
& 31.6{\scriptsize\textcolor{blue}{(+1.2)}}
& 33.5{\scriptsize\textcolor{blue}{(+3.0)}}
& 82.2{\scriptsize\textcolor{blue}{(+1.8)}}
& 33.5{\scriptsize\textcolor{blue}{(+1.3)}}
& 35.2{\scriptsize\textcolor{blue}{(+1.0)}}
& 16.7{\scriptsize\textcolor{blue}{(+0.6)}}
& 23.5{\scriptsize\textcolor{blue}{(+1.1)}}
& 39.7{\scriptsize\textcolor{blue}{(+1.5)}} \\

ClearCLIP {\scriptsize \textcolor{gray}{[ECCV'24]}} \citep{Lan2024ClearCLIP}
& $\times$ & 51.8 & 32.6 & 33.0 & 80.9 & 30.0 & 35.9 & 16.7 & 23.9 & 38.1 \\

ClearCLIP + PTC {\scriptsize [Ours]}
& $\times$
& 51.9{\scriptsize\textcolor{blue}{(+0.1)}}
& 33.4{\scriptsize\textcolor{blue}{(+0.8)}}
& 33.1{\scriptsize\textcolor{blue}{(+0.1)}}
& 82.2{\scriptsize\textcolor{blue}{(+1.3)}}
& 31.7{\scriptsize\textcolor{blue}{(+1.7)}}
& 37.4{\scriptsize\textcolor{blue}{(+1.5)}}
& 18.2{\scriptsize\textcolor{blue}{(+1.5)}}
& 25.1{\scriptsize\textcolor{blue}{(+1.2)}}
& 39.1{\scriptsize\textcolor{blue}{(+1.0)}} \\

NACLIP {\scriptsize \textcolor{gray}{[WACV'25]}} \citep{Hajimiri2025NACLIP}
& $\times$ & 58.8 & 32.2 & 33.2 & 77.1 & 35.5 & 35.2 & 17.4 & 22.9 & 39.0 \\

NACLIP + PTC {\scriptsize [Ours]}
& $\times$
& 62.6{\scriptsize\textcolor{blue}{(+3.8)}}
& 32.9{\scriptsize\textcolor{blue}{(+0.7)}}
& 34.2{\scriptsize\textcolor{blue}{(+1.0)}}
& 82.7{\scriptsize\textcolor{blue}{(+5.6)}}
& 36.8{\scriptsize\textcolor{blue}{(+1.3)}}
& 36.9{\scriptsize\textcolor{blue}{(+1.7)}}
& 18.6{\scriptsize\textcolor{blue}{(+1.2)}}
& 24.7{\scriptsize\textcolor{blue}{(+1.8)}}
& 41.2{\scriptsize\textcolor{blue}{(+2.2)}} \\

ResCLIP {\scriptsize \textcolor{gray}{[CVPR'25]}} \citep{Yang2025ResCLIP}
& $\times$ & 61.1 & 33.5 & 35.0 & 86.0 & 35.9 & 36.8 & 18.0 & 24.7 & 41.4 \\

ResCLIP + PTC {\scriptsize [Ours]}
& $\times$
& 63.9{\scriptsize\textcolor{blue}{(+2.8)}}
& 33.5{\scriptsize\textcolor{blue}{(+0.0)}}
& 35.4{\scriptsize\textcolor{blue}{(+0.4)}}
& 86.9{\scriptsize\textcolor{blue}{(+0.9)}}
& 37.3{\scriptsize\textcolor{blue}{(+1.4)}}
& 37.3{\scriptsize\textcolor{blue}{(+0.5)}}
& 18.3{\scriptsize\textcolor{blue}{(+0.3)}}
& 25.2{\scriptsize\textcolor{blue}{(+0.5)}}
& 42.2{\scriptsize\textcolor{blue}{(+0.8)}} \\

\midrule

ZeroGuideSeg {\scriptsize \textcolor{gray}{[ICCV'23]}}
\citep{Rewatbowornwong2023ZeroGuideSeg}
& $\checkmark$ & - & - & - & 20.1 & - & 19.6 & - & - & - \\

RIM {\scriptsize \textcolor{gray}{[CVPR'24]}} \citep{Wang2024RIM}
& $\checkmark$ & 77.8 & 34.3 & 44.5 & - & - & - & - & - & - \\

FOSSIL {\scriptsize \textcolor{gray}{[WACV'24]}} \citep{Barsellotti2024FOSSIL}
& $\checkmark$ & - & - & - & - & 23.2 & 35.8 & 18.8 & 24.8 & - \\

FreeDA {\scriptsize \textcolor{gray}{[CVPR'24]}} \citep{Barsellotti2024FreeDA}
& $\checkmark$ & 55.4 & 38.3 & 37.4 & 85.6 & 36.7 & 43.1 & 22.4 & 27.8 & 43.3 \\

LLM-Supervision {\scriptsize \textcolor{gray}{[CVPRW'24]}}
\citep{sun2024LLM_Supervision}
& $\checkmark$ & - & - & - & 61.7 & - & 27.8 & - & 29.1 & - \\

LaVG {\scriptsize \textcolor{gray}{[ECCV'24]}} \citep{Kang2025LaVG}
& $\checkmark$ & 62.1 & 31.6 & 34.2 & 82.5 & - & 34.7 & 15.8 & 23.2 & - \\

ProxyCLIP {\scriptsize \textcolor{gray}{[ECCV'24]}} \citep{Lan2024proxyclip}
& $\checkmark$ & 61.3 & 35.3 & 37.5 & 80.3 & 38.1 & 39.1 & 20.2 & 26.5 & 42.3 \\

ProxyCLIP + PTC {\scriptsize [Ours]}
& $\checkmark$
& 67.2{\scriptsize\textcolor{blue}{(+5.9)}}
& 36.2{\scriptsize\textcolor{blue}{(+0.9)}}
& 38.6{\scriptsize\textcolor{blue}{(+1.1)}}
& 82.7{\scriptsize\textcolor{blue}{(+2.4)}}
& 40.0{\scriptsize\textcolor{blue}{(+1.9)}}
& 40.3{\scriptsize\textcolor{blue}{(+1.2)}}
& 20.9{\scriptsize\textcolor{blue}{(+0.7)}}
& 27.5{\scriptsize\textcolor{blue}{(+1.0)}}
& 44.2{\scriptsize\textcolor{blue}{(+1.9)}} \\

CorrCLIP {\scriptsize \textcolor{gray}{[ICCV'25]}} \citep{Zhang2025CorrCLIP}
& $\checkmark$ & 74.8 & 44.2 & 43.7 & 88.8 & 49.4 & 48.8 & 26.9 & 31.6 & 51.0 \\

CorrCLIP + PTC {\scriptsize [Ours]}
& $\checkmark$
& 77.4{\scriptsize\textcolor{blue}{(+2.6)}}
& 44.3{\scriptsize\textcolor{blue}{(+0.1)}}
& 46.2{\scriptsize\textcolor{blue}{(+2.5)}}
& 89.4{\scriptsize\textcolor{blue}{(+0.6)}}
& 49.5{\scriptsize\textcolor{blue}{(+0.1)}}
& 49.2{\scriptsize\textcolor{blue}{(+0.4)}}
& 26.9{\scriptsize\textcolor{blue}{(+0.0)}}
& 32.3{\scriptsize\textcolor{blue}{(+0.7)}}
& 51.9{\scriptsize\textcolor{blue}{(+0.9)}} \\

\bottomrule
\end{tabular}
}
\end{table*}

\paragraph{\textbf{Implementation \& metric}}
Our implementations rely on MMSegmentation \citep{Contributors2020mmseg} and MMEngine \citep{Contributors2022mmengine}.
For image preprocessing, we resize the shorter side of each image to 336 pixels for the PASCAL and COCO datasets, and to 448 pixels for the higher-resolution Cityscapes and ADE20K datasets.
For each input image, PTC extracts reliable visual evidence from overlapping $224\times224$ local crops with a stride of $112\times112$ and aggregates it across all crops to construct category visual prototypes, which are used to calibrate the corresponding text embeddings only once for the entire image.
Final segmentation is performed using a sliding window inference strategy with a $336\times336$ window and a $112\times112$ stride.
Following prior work \citep{Wang2024SCLIP, Lan2024proxyclip}, we obtain class text embeddings using the standard ImageNet prompts \citep{Radford2021CLIP} and the corresponding class names.
We directly evaluate all methods on the validation sets without retraining or fine-tuning.
Segmentation performance is measured using mean intersection over union (mIoU).
No post-processing method, such as DenseCRF \citep{Kraehenbuehl2011DenseCRF} or Pixel-Adaptive Mask Refinement (PAMR) \citep{Araslanov2020PAMR}, is applied. All experiments are conducted on a single NVIDIA RTX 4090 GPU.

\begin{figure*}[t]
\centering
\setlength{\tabcolsep}{0.25mm}
\resizebox{\textwidth}{!}{
{\renewcommand{\arraystretch}{0.35}
\begin{tabular}{c c c c c c c c c c c} 

    \rotatebox[origin=t]{90}{\raisebox{1ex}{\scriptsize \textbf{Image}}} &
    \includegraphics[height=1.355cm]{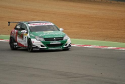} &
    \includegraphics[height=1.355cm]{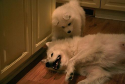} &
    \includegraphics[height=1.355cm]{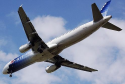} &
    \includegraphics[height=1.355cm]{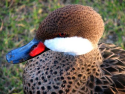} &
    \includegraphics[height=1.355cm]{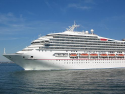} &
    \includegraphics[height=1.355cm]{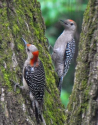} &
    \includegraphics[height=1.355cm]{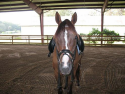} &
    \includegraphics[height=1.355cm]{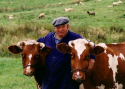} &
    \includegraphics[height=1.355cm]{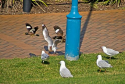} &
    \includegraphics[height=1.355cm]{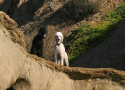} \\
    
    \rotatebox[origin=t]{90}{\raisebox{1ex}{\scriptsize \textbf{ProxyCLIP}}} &
    \includegraphics[height=1.355cm]{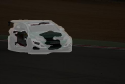} &
    \includegraphics[height=1.355cm]{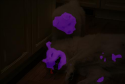} &
    \includegraphics[height=1.355cm]{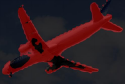} &
    \includegraphics[height=1.355cm]{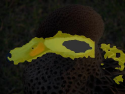} &
    \includegraphics[height=1.355cm]{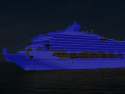} &
    \includegraphics[height=1.355cm]{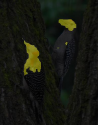} &
    \includegraphics[height=1.355cm]{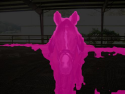} &
    \includegraphics[height=1.355cm]{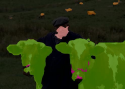} &
    \includegraphics[height=1.355cm]{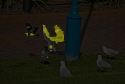} &
    \includegraphics[height=1.355cm]{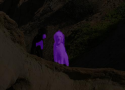} \\
    
    \rotatebox[origin=t]{90}{\raisebox{1ex}{\scriptsize \textbf{+ PTC}}} &
    \includegraphics[height=1.355cm]{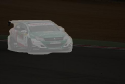} &
    \includegraphics[height=1.355cm]{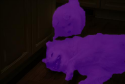} &
    \includegraphics[height=1.355cm]{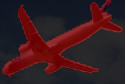} &
    \includegraphics[height=1.355cm]{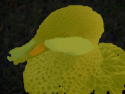} &
    \includegraphics[height=1.355cm]{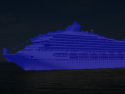} &
    \includegraphics[height=1.355cm]{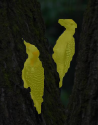} &
    \includegraphics[height=1.355cm]{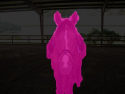} &
    \includegraphics[height=1.355cm]{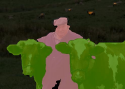} &
    \includegraphics[height=1.355cm]{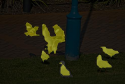} &
    \includegraphics[height=1.355cm]{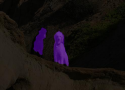} \\

    \rotatebox[origin=t]{90}{\raisebox{1ex}{\scriptsize \textbf{GT}}} &
    \includegraphics[height=1.355cm]{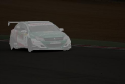} &
    \includegraphics[height=1.355cm]{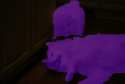} &
    \includegraphics[height=1.355cm]{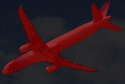} &
    \includegraphics[height=1.355cm]{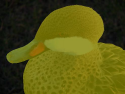} &
    \includegraphics[height=1.355cm]{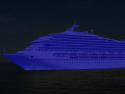} &
    \includegraphics[height=1.355cm]{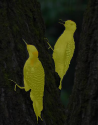} &
    \includegraphics[height=1.355cm]{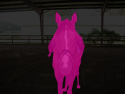} &
    \includegraphics[height=1.355cm]{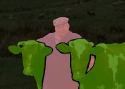} &
    \includegraphics[height=1.355cm]{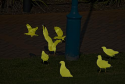} &
    \includegraphics[height=1.355cm]{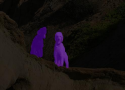} \\

\end{tabular}}}

\caption{Visualization results on the \textbf{PASCAL VOC} dataset. “GT” denotes ground truth.}
\label{fig3}
\end{figure*}

\begin{figure*}[t]
\centering
\setlength{\tabcolsep}{0.25mm}
\resizebox{\textwidth}{!}{
{\renewcommand{\arraystretch}{0.35}
\begin{tabular}{c c c c c c c c c c c}
    
    \rotatebox[origin=t]{90}{\raisebox{1ex}{\scriptsize \textbf{Image}}} &
    \includegraphics[height=1.355cm]{figs/fig4/line1/000000079651.pdf} &
    \includegraphics[height=1.355cm]{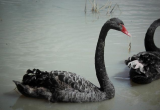} &
    \includegraphics[height=1.355cm]{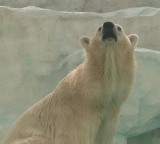} &
    \includegraphics[height=1.355cm]{figs/fig4/line4/000000153217.pdf} &
    \includegraphics[height=1.355cm]{figs/fig4/line5/000000439623.pdf} &
    \includegraphics[height=1.355cm]{figs/fig4/line6/000000183500.pdf} &
    \includegraphics[height=1.355cm]{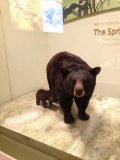} &
    \includegraphics[height=1.355cm]{figs/fig4/line8/000000250758.pdf} &
    \includegraphics[height=1.355cm]{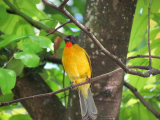} & 
    \includegraphics[height=1.355cm]{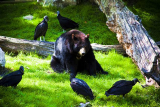} \\
    
    \rotatebox[origin=t]{90}{\raisebox{1ex}{\scriptsize \textbf{ProxyCLIP}}} &
    \includegraphics[height=1.355cm]{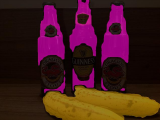} &
    \includegraphics[height=1.355cm]{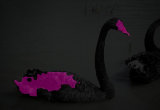} &
    \includegraphics[height=1.355cm]{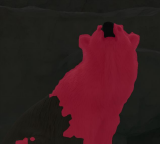} &
    \includegraphics[height=1.355cm]{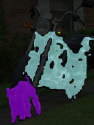} &
    \includegraphics[height=1.355cm]{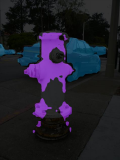} &
    \includegraphics[height=1.355cm]{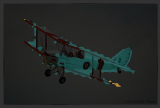} &
    \includegraphics[height=1.355cm]{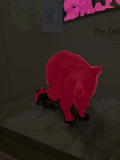} &
    \includegraphics[height=1.355cm]{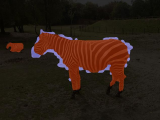} &
    \includegraphics[height=1.355cm]{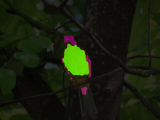} & 
    \includegraphics[height=1.355cm]{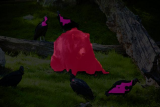} \\
    
    \rotatebox[origin=t]{90}{\raisebox{1ex}{\scriptsize \textbf{+ PTC}}} &
    \includegraphics[height=1.355cm]{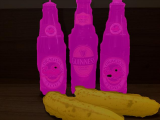} &
    \includegraphics[height=1.355cm]{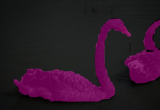} &
    \includegraphics[height=1.355cm]{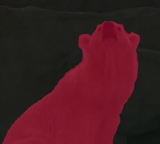} &
    \includegraphics[height=1.355cm]{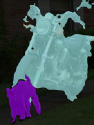} &
    \includegraphics[height=1.355cm]{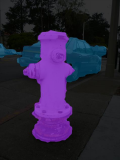} &
    \includegraphics[height=1.355cm]{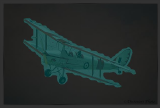} &
    \includegraphics[height=1.355cm]{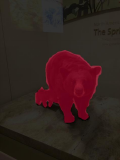} &
    \includegraphics[height=1.355cm]{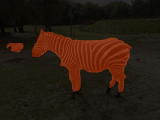} &
    \includegraphics[height=1.355cm]{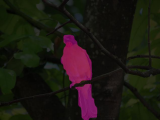} & 
    \includegraphics[height=1.355cm]{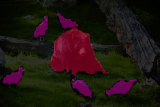} \\

    \rotatebox[origin=t]{90}{\raisebox{1ex}{\scriptsize \textbf{GT}}} &
    \includegraphics[height=1.355cm]{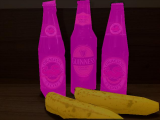} &
    \includegraphics[height=1.355cm]{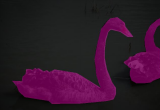} &
    \includegraphics[height=1.355cm]{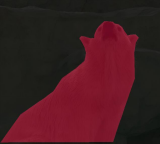} &
    \includegraphics[height=1.355cm]{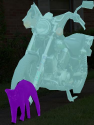} &
    \includegraphics[height=1.355cm]{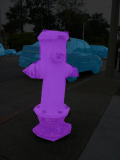} &
    \includegraphics[height=1.355cm]{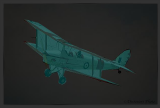} &
    \includegraphics[height=1.355cm]{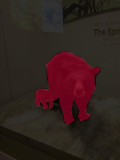} &
    \includegraphics[height=1.355cm]{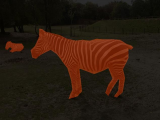} &
    \includegraphics[height=1.355cm]{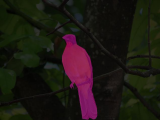} & 
    \includegraphics[height=1.355cm]{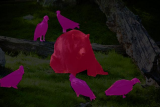} \\

\end{tabular}
}}

\caption{Visualization results on the \textbf{COCO-Object} dataset. “GT” denotes ground truth.}
\label{fig4}
\end{figure*}

\subsection{Comparison Results}
\label{sec:4.2}
\paragraph{\textbf{Compared baselines}}
The compared methods are classified into two groups. 
The first group consists of methods that rely solely on CLIP, including ReCo \citep{Shin2022ReCo}, MaskCLIP \citep{Zhou2022MaskCLIP}, CaR \citep{Sun2024CaR}, GEM \citep{Bousselham2024GEM}, CLIPtrase \citep{Shao2025CLIPTrase}, OVDiff \citep{Karazija2025OVDiff}, SCLIP \citep{Wang2024SCLIP}, ClearCLIP \citep{Lan2024ClearCLIP}, NACLIP \citep{Hajimiri2025NACLIP}, and ResCLIP \citep{Yang2025ResCLIP}. 
The second group comprises methods that incorporate additional models, including ZeroGuideSeg \citep{Rewatbowornwong2023ZeroGuideSeg}, RIM \citep{Wang2024RIM}, FOSSIL \citep{Barsellotti2024FOSSIL}, FreeDA \citep{Barsellotti2024FreeDA}, LLM-Supervision \citep{sun2024LLM_Supervision}, LaVG \citep{Kang2025LaVG}, ProxyCLIP \citep{Lan2024proxyclip}, and CorrCLIP \citep{Zhang2025CorrCLIP}.
Results are presented in \cref{tab:comparison}.

\paragraph{\textbf{Quantitative results}} 
As a plug-and-play text calibration method, PTC can be readily integrated into existing training-free OVSS methods to alleviate the semantic gap between generic category text embeddings and instance-specific visual representations. 
As shown in \cref{tab:comparison}, applying PTC to SCLIP, ClearCLIP, NACLIP, ResCLIP, ProxyCLIP, and CorrCLIP improves their average mIoU by 1.5, 1.0, 2.2, 0.8, 1.9, and 0.9 percentage points, respectively. 
Notably, PTC not only substantially improves baselines with relatively limited initial performance, increasing the average mIoU of NACLIP from 39.0\% to 41.2\%, but also delivers further gains on strong baselines such as CorrCLIP, raising its average mIoU from 51.0\% to 51.9\%. 
These results collectively demonstrate the effectiveness and generalization of PTC.

\paragraph{\textbf{Qualitative results}} 
\cref{fig3,fig4} present qualitative comparisons with and without PTC on the PASCAL VOC and COCO-Object datasets, respectively.
Although ProxyCLIP already localizes object regions reasonably well, its final predictions still rely on generic category text embeddings. Consequently, its recognition tends to focus on highly category-representative local regions and remains susceptible to interference from complex backgrounds and similar classes, leading to incomplete masks, misclassification of local regions, and missed small objects. 
In contrast, PTC yields more complete object coverage, fewer erroneous predictions caused by background interference and class confusion, and improved recognition of small and less salient objects.

On PASCAL VOC, as shown in \cref{fig3}, PTC produces more complete object coverage, more accurate predictions, and better recognition of small objects.
In Examples 1-6, ProxyCLIP often focuses on highly category-representative local regions, such as the heads of dogs and birds, while failing to recognize other parts of the object. After introducing PTC, the model can capture the complete extent of the target object.
As illustrated in Examples 7-8, while background textures, adjacent regions, and occlusions lead ProxyCLIP into local misclassifications, PTC successfully suppresses these interferences, producing more accurate predictions.
In Example 9, PTC successfully captures small objects that are missed by ProxyCLIP.
Moreover, PTC improves the segmentation of a black dog whose dark appearance resembles the shadowed background in Example 10, demonstrating its ability to distinguish foreground from background.

As shown in \cref{fig4}, PTC remains effective on COCO-Object, which contains more categories and more complex scene compositions.
In Examples 1-6, the predictions of ProxyCLIP often leave local regions with relatively weak category cues insufficiently covered.
For example, in Example 1, ProxyCLIP primarily identifies regions with distinctive bottle-like features but misses parts of the bottle whose label texture gives them a different appearance. 
A similar pattern is observed in Examples 2-6.
After calibration with PTC, the text embeddings better match the diverse appearances of instances in the current image, resulting in more complete segmentation.
In Examples 7-9, PTC effectively reduces erroneous predictions caused by complex backgrounds and similar categories, leading to more accurate object segmentation.
In Example 10, PTC successfully captures these challenging small targets, demonstrating its ability to distinguish small objects even in complex scenes.

Overall, PTC can extend the segmentation coverage from highly class-representative local regions to the entire object, reduce erroneous predictions caused by background interference and category confusion, and improve recognition of small and less salient objects.
These results demonstrate that PTC brings the calibrated text embeddings into closer semantic alignment with the category instance appearances in the current image, thereby improving visual-text matching and segmentation performance.

\subsection{Ablation Studies and Analyses}
To systematically validate the effectiveness of PTC, we conduct experiments from three perspectives.
First, we examine the necessity of the core designs of PTC, including margin-based reliability assessment, the hybrid evidence-quantity setting, and evidence-adaptive text calibration.
Second, we analyze the influence of key hyperparameters, including the minimum quantity of visual evidence $K_{\min}$ and the global calibration strength $\mu$, on segmentation performance across datasets.
Finally, we evaluate the applicability of PTC with different VFMs and CLIP backbones.
Unless otherwise specified, all experiments use ProxyCLIP \citep{Lan2024proxyclip} with a ViT-B/16 CLIP backbone and DINO-B/8 as the baseline.

\subsubsection{Ablation on Core Designs of PTC}
\paragraph{\textbf{Effect of Reliability Assessment Criteria}}
To evaluate the proposed margin-based reliability assessment, we compare three criteria: confidence thresholding, confidence-based top-$K_c$ ranking, and margin-based top-$K_c$ ranking.
For both ranking-based criteria, the number of evidence tokens for each category, $K_c$, is determined by the hybrid evidence-quantity strategy of PTC, as defined in \cref{eq:evidence_quantity}. 
Instead, confidence thresholding retains all visual tokens with confidence scores above $\tau$.
The results are reported in \cref{tab:reliability}. 

As shown in \cref{tab:reliability}, confidence thresholding exhibits inconsistent performance across six representative benchmarks. 
At $\tau=0.35$, it improves upon the baseline only on PASCAL VOC21 and COCO-Object, while degrading performance on the remaining datasets, with particularly large drops on Cityscapes, ADE20K-150, and COCO-Stuff. 
In addition, further increasing the confidence threshold does not yield consistent improvements. 
These results indicate that relying solely on a fixed confidence threshold fails to robustly select reliable visual evidence across different categories.
A low threshold tends to include ambiguous tokens, thereby contaminating visual prototypes, whereas a high threshold may leave some categories with insufficient visual evidence, thereby weakening the overall effectiveness of PTC.

Compared to the fixed confidence threshold, both top-$K_c$ selection strategies are more stable and outperform this criterion on all datasets.
Moreover, a closer comparison reveals a slight overall advantage for margin-based top-$K_c$.
Specifically, its performance matches confidence-based top-$K_c$ on PASCAL Context60, COCO-Object, and ADE20K, improves results on PASCAL VOC21 and COCO-Stuff by 0.1\%, and is 0.1\% lower on Cityscapes alone.
Although the performance difference is small, the score margin accounts for the strongest competing category and therefore provides a more direct measure of category-assignment ambiguity than absolute confidence.
Accordingly, we adopt margin-based top-$K_c$ ranking as the default reliability assessment criterion.

\begin{table}[t]
\centering
\caption{\textbf{Ablation studies on different reliability assessment criteria for visual evidence.}}
\label{tab:reliability}
\setlength{\tabcolsep}{7pt}
\renewcommand{\arraystretch}{0.85}
\resizebox{0.8\linewidth}{!}{
\begin{tabular}{lcccccc}
\toprule
Method & VOC21 & Context60 & Object & City & ADE & Stuff \\
\midrule
Baseline               & 61.3 & 35.3 & 37.5 & 38.1 & 20.2 & 26.5 \\
Confidence threshold ($\tau=0.35$) & 62.1 & 35.0 & 38.2 & 36.4 & 12.0 & 17.3 \\
Confidence threshold ($\tau=0.50$) & 55.0 & 34.9 & 37.9 & 29.6 & 12.9 & 17.7 \\
Confidence threshold ($\tau=0.75$) & 37.9 & 35.1 & 37.6 & 11.9 & 17.1 & 21.7 \\
Confidence-based top-$K_c$          & 67.1 & \textbf{36.2} & \textbf{38.6} & \textbf{40.1} & \textbf{20.9} & 27.4 \\
Margin-based top-$K_c$ (Ours)  & \textbf{67.2} & \textbf{36.2} & \textbf{38.6} & 40.0 & \textbf{20.9} & \textbf{27.5} \\
\bottomrule
\end{tabular}
}
\end{table}

\paragraph{\textbf{Effect of Evidence-Quantity Strategies}}
To evaluate the proposed hybrid evidence-quantity strategy, we compare it with two alternatives: fixed-count selection and proportional sampling.
Specifically, the fixed-count strategy uses $K_{\min}$ as the target evidence count for each category, whereas proportional sampling selects $\lfloor \rho N_c \rfloor$ tokens according to the size of the corresponding candidate set.
The hybrid design combines these two components.
For a fair comparison, the proportional and hybrid strategies are evaluated using the same sampling ratios, $\rho \in \{0.1, 0.2, 0.3\}$.
The results are reported in \cref{tab:quantity}.

As shown in \cref{tab:quantity}, the fixed-count strategy already improves upon the baseline, but yields relatively limited gains on Cityscapes, ADE20K-150, and COCO-Stuff, where object scales vary more substantially across categories and scene compositions are more complex.
By assigning the same evidence count regardless of candidate-set size, it provides insufficient coverage for large candidate sets while increasing the inclusion of unreliable tokens in small or noisy candidate sets, thereby limiting the representativeness and reliability of visual prototypes.
By contrast, proportional sampling adjusts the target evidence count according to the size of each category's candidate set, but its performance is highly sensitive to $\rho$. 
Increasing $\rho$ expands evidence coverage but also introduces lower-ranked tokens with smaller score margins; the coverage gain dominates on Cityscapes and COCO-Stuff, whereas the reliability loss dominates on PASCAL VOC21. 
These results show that the size of a candidate set does not reliably indicate the amount of reliable visual evidence it contains, and therefore proportional sampling alone cannot ensure sufficient evidence while avoiding the inclusion of unreliable tokens.

Compared with the fixed-count and proportional sampling strategies, the proposed hybrid strategy delivers the most stable performance across datasets and achieves the best overall results.
Specifically, it improves mIoU over the baseline by 5.9, 0.9, 1.1, 1.9, 0.7, and 1.0 percentage points on PASCAL VOC21, PASCAL Context60, COCO-Object, Cityscapes, ADE20K, and COCO-Stuff, respectively.
For eligible categories with relatively small candidate sets, the minimum-count constraint prevents prototypes from being constructed from too few evidence tokens, whereas for those with large candidate sets, the proportional term scales the evidence count to provide broader coverage of the instance appearance.
By ensuring a minimum level of evidence support while adapting the amount of evidence to variations in object scale, the hybrid strategy provides a more robust approach to visual prototype construction.
Therefore, we adopt it as the default evidence-quantity strategy for PTC.

\begin{table}[t]
\centering
\caption{\textbf{Ablation studies on different visual evidence quantity setting strategies.}}
\label{tab:quantity}
\setlength{\tabcolsep}{7pt}
\renewcommand{\arraystretch}{0.85}
\resizebox{0.8\linewidth}{!}{
\begin{tabular}{lcccccc}
\toprule
Method & VOC21 & Context60 & Object & City & ADE & Stuff \\
\midrule
Baseline                                            & 61.3 & 35.3 & 37.5 & 38.1 & 20.2 & 26.5 \\
Fixed ($K_c=K_{\min}$) & \textbf{67.2} & \textbf{36.2} & 38.5 & 38.5 & 20.7 & 27.1 \\
Proportional ($\rho=0.1$) & 61.2 & 35.4 & 37.5 & 39.0 & 19.7 & 26.9 \\
Proportional ($\rho=0.2$) & 58.7 & 35.7 & 37.6 & 39.6 & 20.5 & 27.3 \\
Proportional ($\rho=0.3$) & 50.5 & 35.9 & 37.6 & 39.6 & 20.7 & 27.3 \\
Hybrid ($\rho=0.1$) & \textbf{67.2} & \textbf{36.2} & \textbf{38.6} & \textbf{40.0 }& \textbf{20.9} & \textbf{27.5} \\
Hybrid ($\rho=0.2$) & 67.1 & \textbf{36.2} & 38.5 & 39.9 & \textbf{20.9} & 27.4 \\
Hybrid ($\rho=0.3$) & 67.1 & \textbf{36.2} & 38.5 & 39.7 & 20.8 & 27.3 \\
\bottomrule
\end{tabular}
}
\end{table}

\paragraph{\textbf{Effect of Text Calibration Schemes}}
To evaluate the proposed evidence-adaptive text calibration, we compare it with uniform calibration, which applies the same strength, $\mu_c=\mu$, to all eligible categories, rather than determining category-specific strengths based on evidence sufficiency.
The results are reported in \cref{tab:calibration}.

Uniform calibration outperforms the baseline on all datasets, demonstrating the effectiveness of calibrating generic category text embeddings with category-specific visual prototypes from the input image.
However, because the amount of available visual evidence and the reliability of the resulting prototypes vary across categories, a uniform calibration strength cannot adequately balance prototype utilization against the preservation of generic category semantics.
Compared with uniform calibration, evidence-adaptive calibration further improves performance across all six datasets, indicating that evidence-dependent calibration strengths enable more effective use of reliable visual information while avoiding excessive modification of original text embeddings for categories with limited evidence.

We further examine the effect of $\lambda$ on evidence-adaptive calibration.
Increasing $\lambda$ from 5 to 10 consistently improves performance across all six datasets, whereas a further increase to 15 yields no additional gains, with decreases in mIoU by 0.1\%, 0.2\%, and 0.1\% on COCO-Object, Cityscapes, and COCO-Stuff, respectively.
According to \cref{eq:calibration}, a larger $\lambda$ assigns a smaller calibration strength for the same amount of visual evidence.
Thus, the calibration is overly conservative when $\lambda$ becomes excessively large, preventing the text embeddings from sufficiently incorporating instance-specific visual information and thereby limiting the effectiveness of PTC.
Overall, $\lambda=10$ achieves the best balance between preserving generic category semantics and incorporating reliable instance-specific visual cues. Therefore, we use $\lambda=10$ as the default value.

\begin{table}[t]
\centering
\caption{\textbf{Ablation studies on different text calibration schemes.}}
\label{tab:calibration}
\setlength{\tabcolsep}{7pt}
\renewcommand{\arraystretch}{0.85}
\resizebox{0.8\linewidth}{!}{
\begin{tabular}{lcccccc}
\toprule
Method  & VOC21 & Context60 & Object & City & ADE & Stuff \\
\midrule
Baseline                      & 61.3 & 35.3 & 37.5 & 38.1 & 20.2 & 26.5 \\
Uniform ($\mu_c=\mu$)              & 64.3 & 36.0 & 38.3 & 39.1 & 20.4 & 26.7 \\
Evidence-adaptive ($\lambda=5$)  & 66.8 & 36.1 & 38.5 & 39.9 & 20.7 & 27.2 \\
Evidence-adaptive ($\lambda=10$) & \textbf{67.2} & \textbf{36.2} & \textbf{38.6} & \textbf{40.0} & \textbf{20.9} & \textbf{27.5} \\
Evidence-adaptive ($\lambda=15$) & 67.2 & 36.2 & 38.5 & 39.8 & 20.9 & 27.4 \\
\bottomrule
\end{tabular}
}
\end{table}

\begin{figure*}[t]
\centering
\captionsetup[subfigure]{font=scriptsize, labelfont=bf, skip=1pt}
\begin{subfigure}[t]{0.245\textwidth}
\centering
\includegraphics[width=\linewidth]{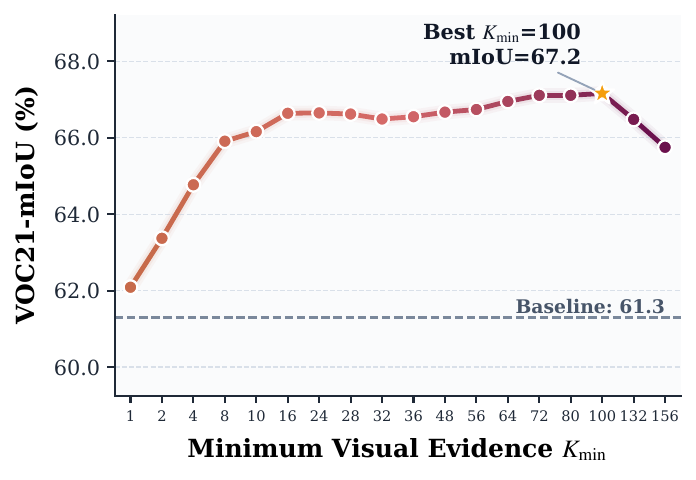}
\caption{PASCAL VOC21}
\label{fig:kmin_voc21}
\end{subfigure}
\hfill
\begin{subfigure}[t]{0.245\textwidth}
\centering
\includegraphics[width=\linewidth]{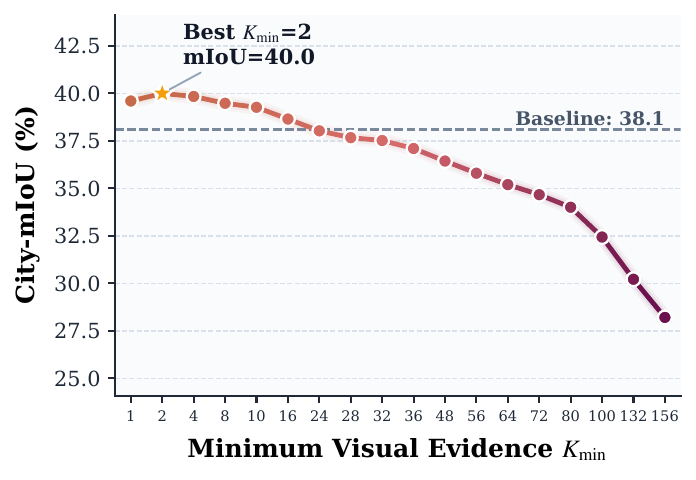}
\caption{Cityscapes}
\label{fig:kmin_city}
\end{subfigure}
\hfill
\begin{subfigure}[t]{0.245\textwidth}
\centering
\includegraphics[width=\linewidth]{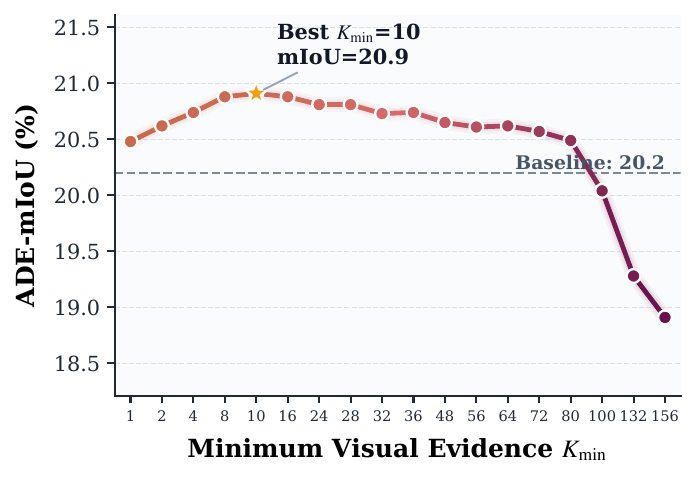}
\caption{ADE20K}
\label{fig:kmin_ade}
\end{subfigure}
\hfill
\begin{subfigure}[t]{0.245\textwidth}
\centering
\includegraphics[width=\linewidth]{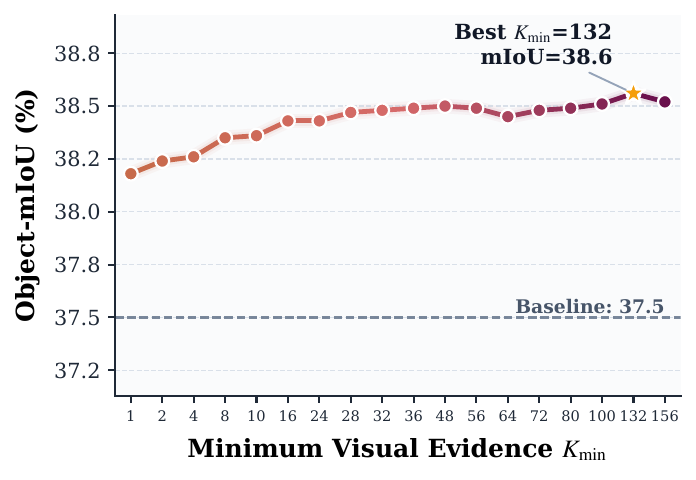}
\caption{COCO-Object}
\label{fig:kmin_object}
\end{subfigure}
    
\vspace{2mm}

\begin{subfigure}[t]{0.245\textwidth}
\centering
\includegraphics[width=\linewidth]{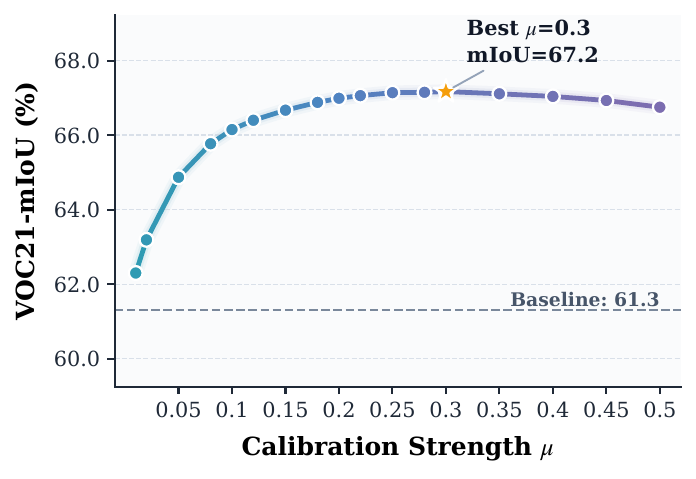}
\caption{PASCAL VOC21}
\label{fig:mu_voc21}
\end{subfigure}
\hfill
\begin{subfigure}[t]{0.245\textwidth}
\centering
\includegraphics[width=\linewidth]{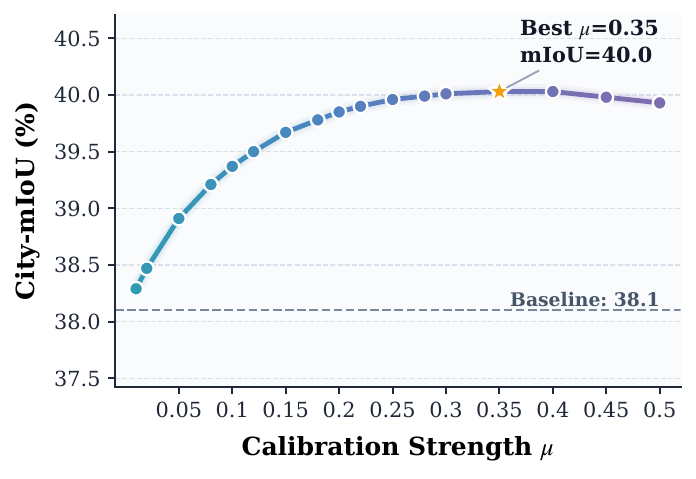}
\caption{Cityscapes}
\label{fig:mu_city}
\end{subfigure}
\hfill
\begin{subfigure}[t]{0.245\textwidth}
\centering
\includegraphics[width=\linewidth]{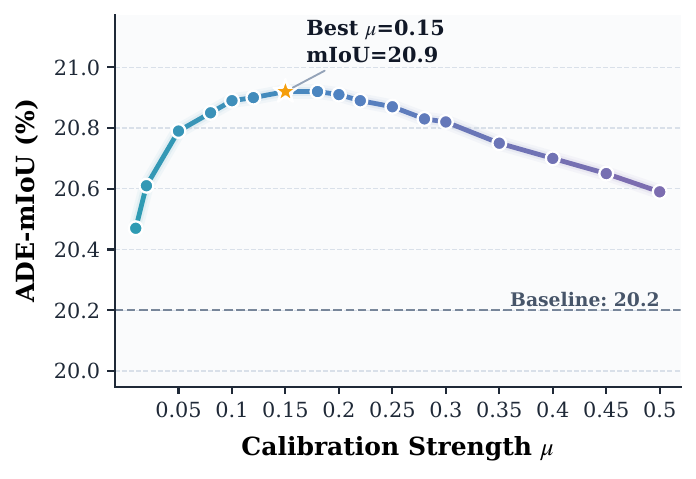}
\caption{ADE20K}
\label{fig:mu_ade}
\end{subfigure}
\hfill
\begin{subfigure}[t]{0.245\textwidth}
\centering
\includegraphics[width=\linewidth]{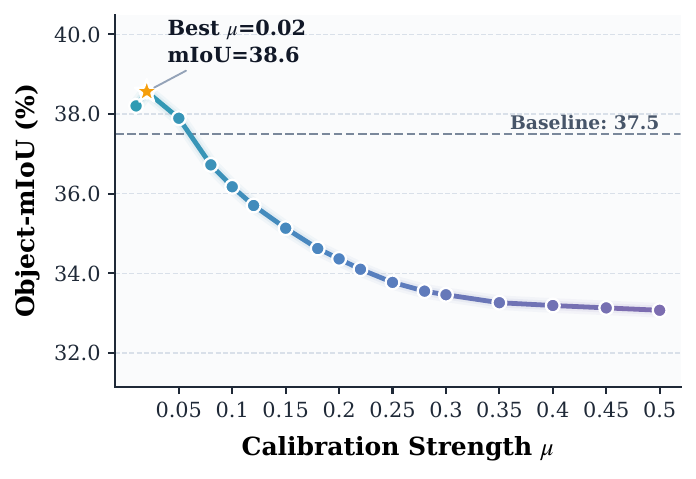}
\caption{COCO-Object}
\label{fig:mu_object}
\end{subfigure}

\caption{\textbf{Sensitivity Analysis of $K_{\min}$ and $\mu$.} \textbf{Top row:} Effect of the minimum quantity of visual evidence ($K_{\min}$) on mIoU across four datasets. \textbf{Bottom row:} Effect of the global calibration strength ($\mu$).}
\label{fig5}
\end{figure*}

\subsubsection{Sensitivity Analysis of $K_{\min}$ and $\mu$}
In this subsection, we analyze the effects of the hyperparameters $K_{\min}$ and $\mu$ on segmentation performance. 
$K_{\min}$ specifies the minimum quantity of visual evidence required for a category to participate in prototype construction and text calibration.
$\mu$ is the global calibration strength, which controls the overall influence of the visual prototype on the original category text embedding.
To examine their individual effects, we vary one parameter at a time while fixing the other at its dataset-specific default value.
The results on four representative datasets are shown in \cref{fig5}. 

Overall, segmentation performance initially improves, then saturates or declines as either $K_{\min}$ or $\mu$ increases.
Within this trend, datasets favor different optimal values due to differences in vocabulary size, the number of categories present in each image, object scales, and scene complexity.

Despite comparably small category vocabularies, PASCAL VOC21 and Cityscapes favor distinct optimal values for $K_{\min}$ and $\mu$ due to differences in scene composition, as well as the number and scale of semantic regions within each image.
As illustrated in \cref{fig5}\subref{fig:kmin_voc21} and \subref{fig:mu_voc21}, PTC achieves its best performance of 67.2\% mIoU on PASCAL VOC21 with $K_{\min}=100$ and $\mu=0.3$.
Images in this dataset generally contain few yet salient foreground objects whose regions are relatively complete and occupy large areas, which can provide a greater amount of reliable visual evidence.
Accordingly, a larger $K_{\min}$ enables more stable visual prototypes to be constructed for well-supported categories while excluding categories supported only by sparse, scattered evidence.
By contrast, PTC achieves its best performance of 40.0\% mIoU on Cityscapes with $K_{\min}=2$ and $\mu=0.35$, as shown in \cref{fig5}\subref{fig:kmin_city} and \subref{fig:mu_city}.
The Cityscapes dataset consists of urban street scenes that typically contain multiple semantic categories with substantial scale differences, some of which are small or elongated.
In this situation, an excessively large $K_{\min}$ prevents categories represented by small or elongated regions from satisfying the minimum evidence requirement, reducing the number of categories eligible for calibration and weakening the gains from PTC.
Notably, both PASCAL VOC21 and Cityscapes achieve better performance with relatively large values of the global calibration strength $\mu$. This indicates that, when reliable visual prototypes can be constructed for eligible categories, appropriately increasing their influence on the text embeddings can improve visual-text matching more effectively.

Moreover, ADE20K and COCO-Object exhibit distinct preferences for $K_{\min}$ and $\mu$, despite comparably large class vocabularies and complex scene content.
For ADE20K, the dataset covers diverse indoor and outdoor scenes, with each image typically containing multiple semantic regions.
PTC achieves its best performance with $K_{\min}=10$ and $\mu=0.15$ on this dataset, as shown in \cref{fig5}\subref{fig:kmin_ade} and \subref{fig:mu_ade}.
This is because a moderate evidence threshold excludes insufficiently supported categories without overly restricting category coverage, while a relatively conservative calibration strength mitigates interference from unreliable visual prototypes in complex scenes.
Compared to the ADE20K dataset, COCO-Object images typically contain fewer categories but often include multiple object instances at different scales, together with complex scene context.
In this setting, a larger $K_{\min}=132$ restricts calibration to categories with sufficient visual evidence, whereas a smaller $\mu=0.02$ limits the influence of potentially erroneous prototypes on the original text embeddings.
In addition, the small $\mu$ reduces the effect of varying $K_{\min}$ on the final predictions, leading to the performance on COCO-Object remaining relatively stable over a broad range of $K_{\min}$ values, as illustrated in \cref{fig5}\subref{fig:kmin_object}.
Conversely, performance on this dataset decreases rapidly as $\mu$ increases, as shown in \cref{fig5}\subref{fig:mu_object}, further indicating that a more conservative calibration magnitude is required in complex scenes.

Overall, differences in vocabulary size, scene composition, and the number and scale of semantic regions lead different datasets to favor different values of $K_{\min}$ and $\mu$.
Based on these results, we adopt dataset-specific configurations for the current baseline.
Moreover, these two parameters are adjusted when PTC is integrated with other baselines, because different baselines produce different initial visual-text matching scores, which affect both the number of candidate tokens assigned to each category and the reliability of the top-ranked tokens selected as visual evidence.
Despite differences in their optimal values, the functional roles and overall sensitivity trends of $K_{\min}$ and $\mu$ remain consistent across datasets and baselines, with all other PTC configurations held constant.

\subsubsection{Ablation on Different Foundation Backbones}
\paragraph{\textbf{Effect of Visual Foundation Models}}
To evaluate the applicability of PTC, we follow ProxyCLIP \citep{Lan2024proxyclip} by replacing its underlying vision foundation model with SAM \citep{kirillov2023sam}, DINO \citep{Caron2021Dino}, and DINOv2 \citep{oquab2023dinov2}, yielding distinct visual features for evaluation. As shown in \cref{tab:vfm}, incorporating PTC consistently improves segmentation performance across all three VFM variants, demonstrating its robustness to diverse visual representations. Specifically, PTC achieves the most notable improvements on PASCAL VOC21, with gains of 2.6\%, 5.9\%, and 3.8\% under the three VFM settings, respectively. This is mainly because PASCAL VOC21 contains fewer categories, and foreground objects in its images are generally larger and more salient, making reliable visual evidence easier to extract. By contrast, ADE20K, COCO-Stuff, and PASCAL Context60 contain more categories and more complex scene content. Category confusion and background interference increase the uncertainty involved in visual prototype construction, resulting in more modest performance gains. Nevertheless, PTC still yields gains overall on these challenging datasets, demonstrating its effectiveness in improving the accuracy of visual-text alignment. These results collectively validate the generalization of PTC with different vision foundation models.
\begin{table}[t]
\centering
\caption{\textbf{Ablation studies with different visual foundation models.}}
\vspace{-10pt}
\label{tab:vfm}
\setlength{\tabcolsep}{7pt}
\renewcommand{\arraystretch}{0.85}
\resizebox{0.8\linewidth}{!}{
\begin{tabular}{llcccccc}
\toprule
VFMs & Method & VOC21 & Context60 & Object & City & ADE & Stuff \\
\midrule
\multirow{2}{*}{\makecell[l]{SAM\\[-3pt](B/16)}}
& Baseline & 59.3 & 33.6 & 35.4 & 37.0 & \textbf{19.1} & 25.0 \\
& +PTC     & \textbf{61.9} & \textbf{34.4} & \textbf{36.2} & \textbf{38.0} & 18.9 & \textbf{25.6} \\
\midrule
\multirow{2}{*}{\makecell[l]{DINO\\[-3pt](B/8)}}
& Baseline & 61.3 & 35.3 & 37.5 & 38.1 & 20.2 & 26.5 \\
& +PTC     & \textbf{67.2} & \textbf{36.2} & \textbf{38.6} & \textbf{40.0} & \textbf{20.9} & \textbf{27.5} \\
\midrule
\multirow{2}{*}{\makecell[l]{DINOv2\\[-3pt](B/14)}}
& Baseline & 58.6 & 33.8 & 37.4 & 33.9 & 19.7 & 25.4 \\
& +PTC     & \textbf{62.4} & \textbf{34.2} & \textbf{37.7} & \textbf{35.2} & \textbf{20.1} & \textbf{26.1} \\
\bottomrule
\end{tabular}
}
\end{table}

\begin{table}[t]
\centering
\caption{\textbf{Ablation studies with different CLIP backbones.}}
\vspace{-10pt}
\label{tab:clip_backbone}
\setlength{\tabcolsep}{7pt}
\renewcommand{\arraystretch}{0.85}
\resizebox{0.8\linewidth}{!}{
\begin{tabular}{llcccccc}
\toprule
CLIP & Method & VOC21 & Context60 & Object & City & ADE & Stuff \\
\midrule
\multirow{2}{*}{ViT-B/16}
& Baseline & 61.3 & 35.3 & 37.5 & 38.1 & 20.2 & 26.5 \\[-1pt]
& +PTC     & \textbf{67.2} & \textbf{36.2} & \textbf{38.6} & \textbf{40.0} & \textbf{20.9} & \textbf{27.5} \\
\midrule
\multirow{2}{*}{ViT-L/14}
& Baseline & 60.6 & 34.5 & 39.2 & 40.1 & 22.6 & 25.6 \\[-1pt]
& +PTC     & \textbf{65.3} & \textbf{34.8} & \textbf{38.8} & \textbf{41.3} & \textbf{23.0} & \textbf{26.7} \\
\midrule
\multirow{2}{*}{ViT-H/14}
& Baseline & 65.0 & 35.4 & 38.6 & 42.0 & 24.2 & 26.8 \\[-1pt]
& +PTC     & \textbf{67.7} & \textbf{36.4} & \textbf{39.0} & \textbf{42.2} & \textbf{24.5} & \textbf{27.2} \\
\bottomrule
\end{tabular}
}
\end{table}

\paragraph{\textbf{Effect of CLIP Backbones}}
To verify the applicability of PTC, we further adopt different CLIP backbones on the DINO-B/8-based ProxyCLIP baseline. As shown in \cref{tab:clip_backbone}, PTC produces overall performance gains with ViT-B/16, ViT-L/14, and ViT-H/14, demonstrating that the proposed text calibration strategy can be effectively applied across different CLIP backbones. Across the three CLIP backbones, PTC continues to produce the most significant improvements on PASCAL VOC21, increasing mIoU by 5.9\%, 4.7\%, and 2.7\%, respectively. This may be because the primary objects in this dataset are typically well defined, making reliable category-specific visual prototypes easier to construct. As the backbone size increases, the average gain from PTC gradually narrows, which may be because stronger CLIP representations leave less room for further improvement. Nevertheless, PTC achieves overall performance improvements on all three backbones. These findings indicate that the semantic gap between generic category text embeddings and category-specific visual representations in the current image does not entirely disappear as the representations become stronger, and that PTC remains effective in mitigating this gap. Notably, when ViT-L/14 is used as the backbone, PTC produces a slight decrease on COCO-Object. This result indicates that in scenes with a large target vocabulary and strong background interference, the reliability of category-specific visual prototypes is more susceptible to these factors, thereby weakening the gains from text calibration. Despite this, the overall results demonstrate that PTC remains broadly applicable across different CLIP backbones.

\section{Conclusion}
\label{sec5}
In this work, we investigate an underexplored limitation in training-free OVSS: the semantic gap between generic category text embeddings and instance-specific visual representations, which hinders segmentation performance.
\textcolor{black}{To address this issue, we propose PTC, which extracts reliable visual evidence to construct category-specific prototypes and subsequently uses these prototypes to adaptively calibrate the corresponding text embeddings.}
In this way, the calibrated text embeddings incorporate visual cues of the corresponding category instances in the input while preserving generic category semantics, thereby improving the accuracy of visual-text alignment.
PTC does not alter the feature extraction process of existing methods and requires neither additional training nor external models, thereby allowing it to serve as a plug-and-play module for various training-free OVSS methods.
Experiments across eight benchmarks show that PTC significantly improves segmentation performance, confirming that establishing a perceptual link between generic category text embeddings and instance-specific visual representations effectively alleviates their semantic gap.
Nevertheless, the performance of PTC remains dependent on the quality of the constructed visual prototypes. Future work will investigate more reliable prototype construction in complex scenes and further improve the robustness of text calibration.

\bibliographystyle{elsarticle-num-etal3}   
\bibliography{my_refs}    
\end{document}